\documentclass[11pt]{article}

\usepackage[]{acl}
\usepackage[most]{tcolorbox}
\usepackage{times}
\usepackage{latexsym}
\usepackage{xspace}

\def\lays{\textit{LaySPA}\@\xspace}

\def\eg{\textit{eg}\@\xspace}
\usepackage[T1]{fontenc}
\usepackage[utf8]{inputenc}

\usepackage{microtype}
\usepackage{inconsolata}
\usepackage{amssymb}
\usepackage{dsfont}
\usepackage{amsmath}
\usepackage{booktabs}
\usepackage{multirow}
\usepackage[normalem]{ulem}
\useunder{\uline}{\ul}{}
\usepackage{amssymb}
\usepackage{fvextra} 
\usepackage{bbm}
\usepackage{graphicx}
\usepackage{colortbl}
\usepackage{amsthm}
\usepackage{mathrsfs}
\usepackage{makecell}
\usepackage[linesnumbered,ruled,vlined]{algorithm2e}
\usepackage{amsmath}
\usepackage[dvipsnames]{xcolor}
\usepackage{svg}
\usepackage{caption}
\usepackage{comment}
\usepackage{times}
\usepackage{latexsym}
\usepackage{multirow}
\usepackage{array}
\usepackage{subcaption}
\usepackage{amssymb}
\usepackage{pifont}
\usepackage{graphicx}
\usepackage{svg}
\svgsetup{inkscapelatex=false}
\usepackage{enumitem}
\usepackage{graphicx} 
\setlist[itemize]{itemsep=0pt, parsep=0pt, topsep=0pt, partopsep=0pt}
\usepackage{booktabs}
\usepackage{multirow}
\usepackage{graphicx} % 

\usepackage{xspace}

\def\lays{\textit{LaySPA}\@\xspace}

\def\eg{\textit{e.g.}\@\xspace}
\usepackage[utf8]{inputenc}
\usepackage[table]{xcolor} % in preamble

\definecolor{color1}{HTML}{AEC7E8} % light blue
\definecolor{color2}{HTML}{FFBB78} % orange
\definecolor{color3}{HTML}{98DF8A} % green
\definecolor{color4}{HTML}{FF9896} % red/p
\usepackage{microtype}

\usepackage{inconsolata}

\usepackage{graphicx}

\title{From Pixels to Policies: Reinforcing Spatial Reasoning in Language Models for Content-Aware Layout Design}

\author{
\textbf{Sha Li}\textsuperscript{1},
\textbf{Stefano Petrangeli}\textsuperscript{2},
\textbf{Yu Shen}\textsuperscript{2},
\textbf{Xiang Chen}\textsuperscript{2} \\
\textsuperscript{1}Virginia Tech \\
\textsuperscript{2}Adobe Research
}

\begin{document}
\maketitle
\begin{abstract}
We introduce \lays, a reinforcement learning framework that equips large language models (LLMs) with explicit and interpretable spatial reasoning for content-aware graphic layout design. \lays addresses two key challenges: LLMs’ limited spatial reasoning and the lack of opacity in design decision making. Instead of operating at the pixel level, we reformulate layout design as a policy learning problem over a structured textual spatial environment that explicitly encodes canvas geometry, element attributes, and inter-element relationships. \lays produces dual-level outputs comprising interpretable reasoning traces and structured layout specifications, enabling transparent and controllable design decision making. Layout design policy is optimized via a multi-objective spatial critique that decomposes layout quality into geometric validity, relational coherence, and aesthetic consistency, and is trained using relative group optimization to stabilize learning in open-ended design spaces. Experiments demonstrate that \lays improves structural validity and visual quality, outperforming larger proprietary LLMs and achieving performance comparable to specialized SOTA layout generators while requiring fewer annotated samples and reduced latency.
\end{abstract}

\section{Introduction}
\vspace{-0.2em}
\label{sec:intro}
Large Language Models (LLMs) have demonstrated strong capabilities in structured reasoning \citep{zhang2025ratt}, planning \citep{ferrag2025llm} and decision making \citep{zhai2025enhancing}. These advances extend beyond natural language understanding and generation to domains requiring multi-step reasoning \citep{plaat2025multi}, interaction \citep{chen2025analyzing}, and symbolic manipulation, positioning LLMs as a promising substrate for design-oriented tasks that demand explicit relational and structural reasoning \citep{wu2025large}. Within this context, \textit{automatic content-aware graphic layout design} \citep{zheng2019content} emerges as a compelling testbed for LLM-centric spatial reasoning, with applications spanning advertising \citep{hsu2023posterlayout}, web interfaces \citep{wang-etal-2025-banneragency}, and poster design \citep{chai2023two}, all of which require coordinated placement of multiple elements under intertwined semantic, geometric, and aesthetic constraints. 

\begin{figure}[h!]
  \centering
  \subfloat[\small Input canvas\label{fig:canvas}]{
    \includegraphics[width=0.15\textwidth]{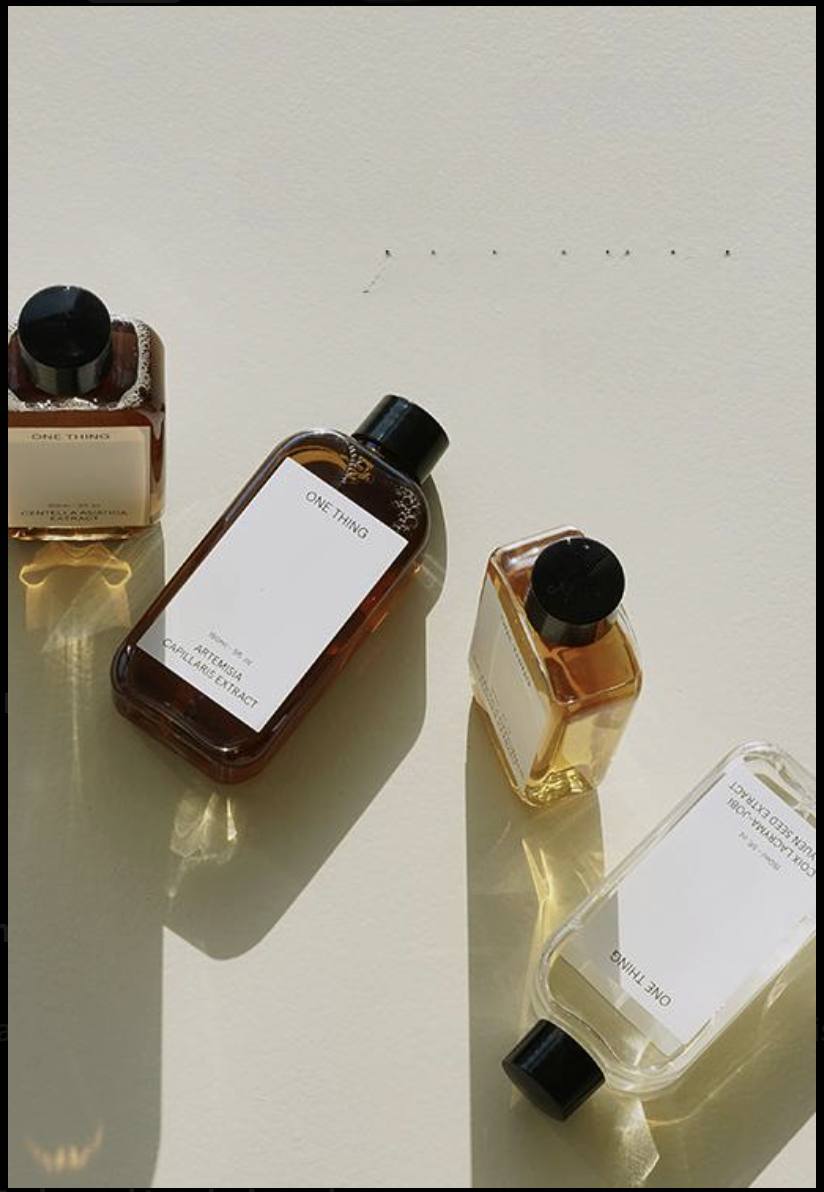}
  }
% optional spacing between figures
  \subfloat[\small GPT-5 output \label{fig:ex_gpt5}]{
    \includegraphics[width=0.15\textwidth]{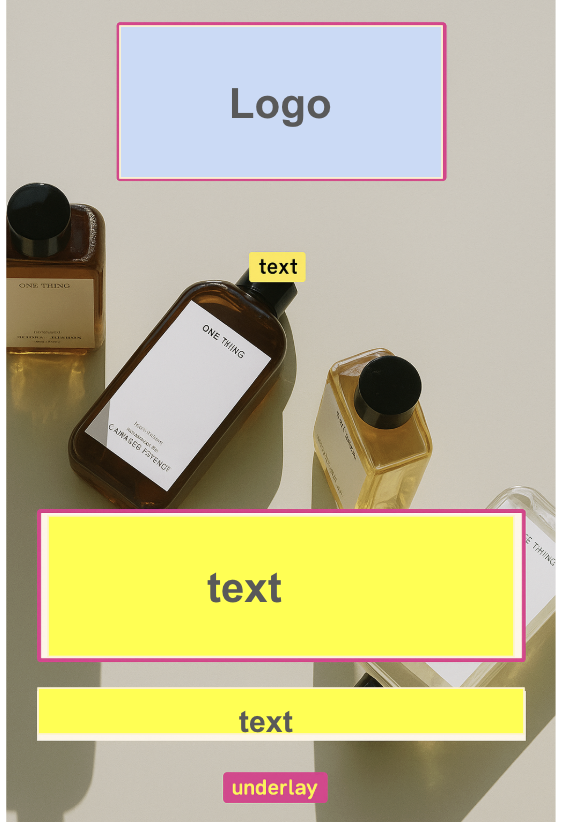}
  }
    \subfloat[\small Human-designed \label{fig:ex_gt}]{
    \includegraphics[width=0.15\textwidth]{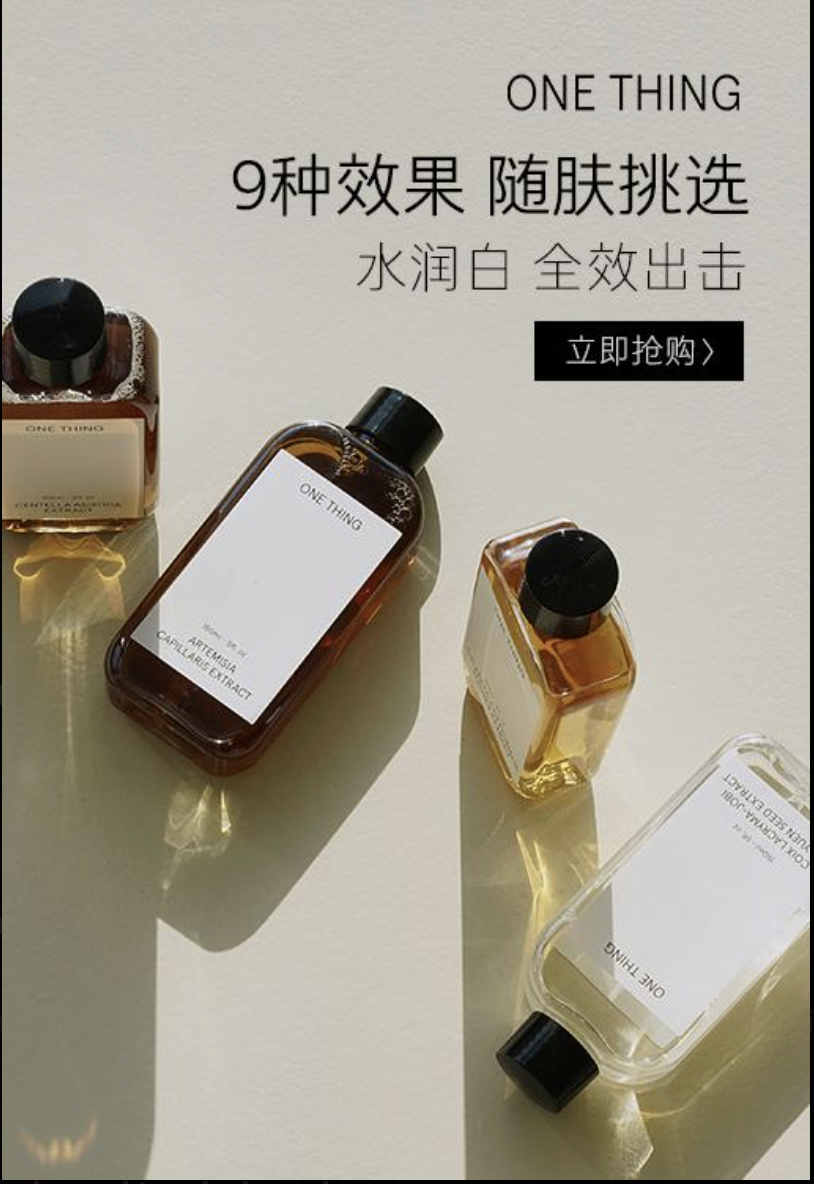}
  }
\vspace{-1.6em}
  \caption{Comparison between a human-designed poster and a GPT-5–generated layout, illustrating common spatial reasoning failures in LLM-based layout generation.}
\vspace{-1.em}
  \label{fig:intro_example}
\end{figure}

Prior approaches typically formulate layout design as coordinate regression or image-conditioned synthesis, using diffusion models \citep{chai2023layoutdm, zhang2023layoutdiffusion} or autoregressive generators \citep{horita2024retrieval}. While effective at producing visually plausible layouts, these methods rely heavily on large annotated datasets and offer limited interpretability: \textit{spatial decisions are entangled with perception in dense latent representations, often prioritizing appearance over structural principles}. In contrast, layout design requires explicit global spatial coherence, multi-object relational reasoning and semantic–functional coupling (\eg, hierarchy, alignment, decoration), and generalizability to novel element configurations or constraints, which distinguishes it from pixel-level image generation and expose the limitations of purely perceptual modeling.

LLMs provide a complementary perspective as they encode prior knowledge of design principles \citep{feng2023layoutgpt}, support symbolic reasoning \citep{wu2023symbol} and produce interpretable outputs. Early attempts apply LLMs to layout generation via structured serializations using in-context learning \citep{horita2024retrieval}, chain-of-thought reasoning \citep{wei2022chain}, or prompt engineering \citep{lin2023layoutprompter}. However, these approaches treat spatial reasoning as an \textit{emergent byproduct} instead of an explicit modeling objective: they neither represent spatial constraints explicitly nor enforce geometric or relational consistency at the element or canvas level. Multimodal extensions using visual–language models \citep{ranasinghe2024learning, cheng2024spatialrgpt} improve visual grounding but still struggle with structural consistency and geometric plausibility. As illustrated in Fig.~\ref{fig:intro_example}, GPT-5 exhibits systematic layout errors including misalignment, inconsistent sizing and violated semantic relations (\eg, underlay–text), underscoring that effective layout design requires explicit reasoning over spatial constraints and relational geometry, not merely visual grounding. Moreover, multimodal architectures often rely on inductive biases to implicitly encode geometry. A text-only formulation removes these shortcuts, making spatial reasoning an explicit and learnable capability.

These observations motivates a new research question: \textbf{\textit{How can an LLM learn robust and generalizable spatial reasoning for layout design from textual serialization without pixel-level supervision?}} Addressing this question requires moving beyond prompt-based generation and multimodal shortcuts. However, LLMs are known to struggle with spatial reasoning, particularly in tasks involving relative geometry and multi-object relationships \citep{bang-etal-2023-multitask, zha2025enable}. In layout design, these limitations are exacerbated by (1) \textit{the lack of intrinsic geometric reasoning capability} for alignment, containment and hierarchy, and (2) \textit{open-ended design spaces with sparse supervision}, where multiple layouts may be equally valid and paired annotations are scarce.

To address these challenges, we propose \lays, a framework that equips LLMs with explicit spatial reasoning for content-aware layout design. We formulate layout generation as a \textit{policy learning} problem in a spatially grounded environment, where an LLM optimizes design decisions through multi-objective spatial critique. By decoupling spatial reasoning from pixel-level generation, \lays enables interpretable decision making, geometry-driven generalization, and direct optimization of spatial competence via reward design

Our contributions are threefold:
(1) We reformulate content-aware layout design as a policy learning problem, enabling explicit and interpretable spatial decision making beyond pixel-level generation. (2) We reinforce LLMs’ spatial reasoning over serialized canvases via multi-objective critiques that encode geometric, relational, and aesthetic constraints as learnable objectives. (3) We empirically demonstrate that \lays learns robust spatial reasoning, outperforming larger proprietary LLMs and most visual-based layout generators, while elucidating the complementary strengths and limitations of language-only and vision-based approaches.

\section{Method}
\label{sec:method}
\vspace{-0.2em}
\subsection{Problem Formulation}
\vspace{-0.2em}
We formulate content-aware layout design as a \textit{policy learning} problem, where an LLM learns a design policy $\pi_{\theta}$ that performs explicit spatial reasoning over a structured textual canvas specification $\mathcal{S}$, generating layouts $\mathcal{L} = \pi_{\theta}(\cdot \mid \mathcal{S})$ that are structurally coherent and visually appealing, without pixel-level supervision and with spatial reasoning as the primary learning signal.

\begin{figure*}[htb]
    \centering
    \includegraphics[width=1.0\textwidth]{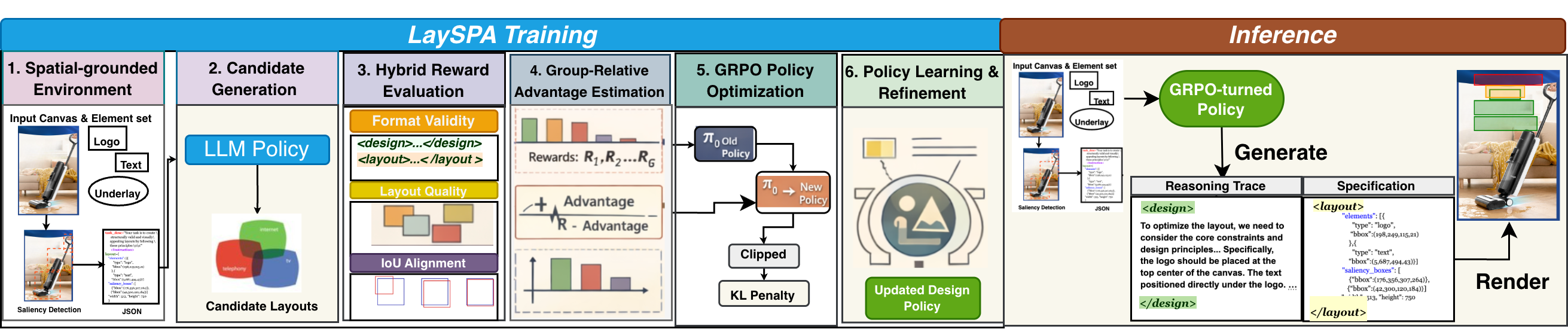}
    \vspace{-1.8em}
    \caption{The framework and workflow of \lays.}
    \label{fig:overview}
    \vspace{-1.5em}
\end{figure*}
\vspace{-0.5em}
\subsection{\lays}
\vspace{-0.4em}
We propose \lays (Fig.~\ref{fig:overview}), a reinforcement learning (\S\ref{sec:rl}) framework that grounds the canvas as a structured spatial environment (\S\ref{sec:env}), produces interpretable layout actions (\S\ref{sec:duallevel}), and optimizes the LLM’s design policy via a multi-objective critique (\S\ref{sec:critique}).

\vspace{-0.4em}
\subsubsection{Spatially-Grounded Environment}
\label{sec:env}
\lays represents each canvas as a spatially grounded environment that encodes normalized coordinates, semantic elements (\eg, text, logos, underlays), and salient regions to be avoided. This structured abstraction exposes geometric and relational constraints directly to the LLM. Formally, a layout $\mathcal{L}$ comprises a set of elements $E$ and saliency regions $S$, where each item is defined by top-left coordinates $(x_i, y_i)$, width and height $(w_i, h_i)$, and category $c_i$. Conditioned on element semantics and the saliency context, the LLM infers the geometric attributes $(x_i, y_i, w_i, h_i)$ for all elements. Layouts are serialized in JSON.

\vspace{-0.5em}
\subsubsection{Dual-Level Design Artifacts} 
\label{sec:duallevel}
\vspace{-0.2em}
Given a canvas specification $\mathcal{S}=\{E,S\}$, the LLM policy $\pi_\theta$ generates layouts ${\ell}_{g=1}^{G}$, each comprising: (i) an explicit \textit{design reasoning trace} that articulates the model’s spatial reasoning and design logic, and (ii) a \textit{structured layout specification} that encodes precise geometric configurations of all elements. These outputs are specialized in $\langle design\rangle$ and $\langle layout\rangle$ blocks, respectively. This dual-level representation decouples relational reasoning from geometric instantiation, enabling interpretability, error diagnosis, and reward computation for downstream evaluation and optimization.
\vspace{-0.2em}
\subsubsection{Multi-Objective Spatial Critique} 

\label{sec:critique}
To guide policy learning, we introduce a multi-objective spatial critique that evaluates candidate layouts and assigns normalized scalar rewards capturing geometric validity, structural coherence and visual aesthetics. This structured feedback enables the policy to differentiate fine-grained spatial trade-offs and converge toward globally coherent and visually balanced layouts.
\vspace{-0.5em}
\paragraph{1) Format Correctness Reward $R_{\text{format}}$} enforces strict compliance with the required \textit{design–layout} output schema, ensuring that each output is structurally complete and parsable. Specifically, outputs must contain a $\langle\textit{design}\rangle$ block that articulates design rationale and a $\langle\textit{layout}\rangle$ block that specifies element geometries in valid JSON. The reward is assigned hierarchically: 0.1 if either block is missing, 0.2 if the JSON is unparsable, 0.5 if the JSON is valid but element types or counts mismatch the input, and 1.0 if both blocks are present, the JSON is valid, and all element types and counts match the specification. $R_{\text{format}}$ mitigates hallucinated or malformed outputs and guarantees compatibility with downstream geometric evaluation. 
\vspace{-1em}
\paragraph{2) Layout Quality Reward $R_{\text{quality}}$} evaluates layouts based on geometric constraints and design principles using five normalized sub-metrics.

\noindent\textbf{2.1) Inverse Collision Rate} $S_{\text{icr}}$ penalizes unintended overlaps between incompatible elements (\eg, text, logos, embellishments and salient regions), while permitting intended overlaps between underlays and their associated text. Each element is represented as an axis-aligned bounding box $b_i = (x_i, y_i, w_i, h_i)$. For any pair $(b_i, b_j)$, the pairwise inverse collision score is defined as:

\vspace{-1.2em}
{\small
\begin{equation}
\label{eq:icr}
S_{\text{icr}}(b_i, b_j)
= 1 - \frac{\mathrm{Area}(b_i \cap b_j)}
{\mathrm{Area}(b_i) + \mathrm{Area}(b_j) - \mathrm{Area}(b_i \cap b_j)}
\end{equation}
}where $\mathrm{Area}(\cdot)$ denotes the area of a bounding box. Higher scores correspond to layouts with minimal unintended overlap.

\noindent\textbf{2.2) Alignment Score} $S_{\text{align}}$ quantifies layout regularity by measuring (i) global centering relative to the canvas, and (ii) mutual alignment among elements. Well-aligned layouts exhibit small deviations from the canvas center and low dispersion of element centers. Consider a layout with $N$ elements, the center of element $e_{i}$ {\small$\boldsymbol{\mu}_i = \big(x_i + \tfrac{w_i}{2}, \; y_i + \tfrac{h_i}{2}\big)$}, 
and the canvas center {\small$\boldsymbol{\mu}_{canvas} = \big(\tfrac{w_{canvas}}{2}, \; \tfrac{h_{canvas}}{2}\big)$}. Then we measure:

\noindent(i) Element-to-canvas alignment: valuating global centering by the normalized average Euclidean distance between element and canvas center:

\vspace{-1em}
{\small\begin{equation}
\vspace{-0.7em}
A_{\text{ele-canvas}} = 1 - \frac{1}{N} \sum_{i=1}^{N} 
\frac{\| \mu_i - \mu_{\text{canvas}} \|_2}{D_{\max}}
\end{equation}} where {\small$D_{\max} = \sqrt{w_c^2 + h_c^2}$} ensures scale invariance.

\noindent(ii) Element-to-element alignment: capturing mutual alignment by penalizing the variance of element centers along the horizontal and vertical axes: 

\vspace{-0.6em}
{\small
\begin{equation}
A_{\text{ele-ele}}
= 1 - \frac{1}{2} \Big(
\mathrm{Var}(\mu_{i,x}) + \mathrm{Var}(\mu_{i,y})
\Big)
\vspace{-0.6em}
\end{equation}
}
where $\text{Var}(\cdot)$ is computed across all element centers along each axis. Lower variance corresponds to higher alignment.

\noindent The final alignment score is computed as:

\vspace{-0.8em}
{\small
\begin{equation}
S_{\text{align}} = \alpha \cdot\, A_{\text{ele-canvas}} + (1-\alpha) \cdot A_{\text{ele-ele}}
\vspace{-0.6em}
\end{equation}
}where $\alpha \in [0,1]$ balances between global centering and inter-element alignment.

\noindent\textbf{2.3) Distribution Score} $S_{\text{dist}}$  measures how evenly elements are arranged on the canvas by jointly evaluating spatial dispersion and area coverage. It combines (i) a normalized spread variance that encourages broad spatial distribution, and (ii) a grid coverage score that promotes utilization of the canvas while discouraging element clustering.

The normalized spread variance is defined as:
\vspace{-0.5em}
\begin{equation}
    V_{\text{spread}}
= \frac{1}{N} \sum_{i=1}^{N}
\frac{\lVert \boldsymbol{\mu}_i - \bar{\boldsymbol{\mu}} \rVert_2^2}{D_{\max}^2}
\vspace{-0.5em}
\end{equation}where $\bar{\boldsymbol{\mu}} = (\bar{x}, \bar{y}) = \tfrac{1}{N} \sum_{i=1}^{N} \boldsymbol{\mu}_i$ is the average center of all elements. Higher values indicate broader spatial dispersion. 

Grid coverage measures area utilization by partitioning the canvas into a $3 \times 3$ grid. Let $G_{ij} \in \{0,1\}$ indicate whether grid cell $(i,j)$ contains at least one element center. The coverage score is:

\vspace{-0.6em}
{\small\begin{equation}
    V_{\text{grid}}
= \frac{1}{9} \sum_{i=1}^{3} \sum_{j=1}^{3} G_{ij}
\vspace{-0.6em}
\end{equation}}
\noindent The final distribution score is computed by:
\vspace{-1em}
\begin{equation}
S_{\text{dist}}
= \frac{1}{2} \big( V_{\text{spread}} + V_{\text{grid}} \big)
\vspace{-0.6em}
\end{equation}

\noindent\textbf{2.4) Spacing Consistency Score} $S_{\text{spacing}}$ quantifies vertical rhythm by computing the normalized variance of distances between adjacent element centers. Elements are first sorted by vertical positions, then the distances between adjacent centers are computed. The variance of these spacings is normalized by the mean spacing and canvas height. Higher scores correspond to more even spacing. Formally, let $c_i^y = y_i + h_i/2$ denote the vertical center of element $e_{i}$. After sorting elements by
$c_i^y$, the adjacent spacings are $s_i = c_{(i+1)}^y - c_{(i)}^y$. The spacing consistency is defined as:

\vspace{-0.6em}
{\small\begin{equation}
S_{\text{spacing}} = 1 - \frac{\text{Var}(s_1, \dots, s_{N-1})}{\bar{s}^2}
\vspace{-0.3em}
\end{equation}}where  $\bar{s}$ is the average spacing. Larger values indicate more uniform vertical spacing.

\noindent\textbf{2.5) Underlay-Text Constraint Score} $S_{\text{underlay}}$ enforces semantic consistency by aligning each underlay with exactly one corresponding text element. Layouts are penalized if multiple text elements overlap a single underlay or if the underlay’s position and size deviate from its associated text. Exact alignment receives a score of 1, partial matches receives intermediate scores, and incorrect pairings are scored 0. This mechanism promotes readability and preserves intended functional relationships between decorative and content elements.

Finally, these sub-rewards are aggregated into a weighted layout quality score. Appendix.~\ref{app:visreward} Figure~\ref{fig:textimagegrid} illustrates how these scores guide the model in balancing structural feasibility and visual appeal in content-aware layouts.

\vspace{-0.3em}
\paragraph{3) IoU Matching Reward $R_{\text{IoU}}$} provides external supervision by measuring the geometric agreement between generated layouts and human-designed references. For each element, IoU is computed as the ratio of the intersection to the union area of the predicted and ground-truth bounding boxes. Unlike intrinsic metrics such as collision or alignment, which assess internal structural consistency, $R_{\text{IoU}}$ anchors learning to human layout exemplars that implicitly encode balance, proportion, and functional grouping. This signal guides the policy toward realistic spatial organization and visually coherent designs. 
\vspace{-0.6em}
\paragraph{4) Final Hybrid Reward} for each generated layout is a weighted combination of:
\vspace{-0.6em}
\begin{equation}
\label{eq:reward}
R(L_{i})=\lambda_{f}R_{format} + \lambda_{q}R_{quality} + \lambda_{u}R_{IoU}
\vspace{-0.5em}
\end{equation}

\subsection{Training Procedure} 
\vspace{-0.3em}
\label{sec:rl}
We update the design policy using Group Relative Policy Optimization (GRPO) \citep{shao2024deepseekmath}, which uses group-wise comparisons as relative baselines. At each step, the old policy $\pi_{\theta_{\text{old}}}$ samples a group of candidate layouts $G$, each scored $R(\ell_i)$ by the hybrid reward in Eq.~\ref{eq:reward}. Relative advantages are computed by normalizing rewards within the group {\small$A(\ell_{i})=\frac{R(\ell_{i})-mean_{\ell_i \in G}R(\ell_{i})}{std_{\ell_i \in G}R(\ell_{i})}$}, providing stable learning signals without an explicit critic. The policy $\pi_\theta$ is then optimized using the clipped surrogate objective:
\vspace{-0.5em}
\begin{equation}
\label{eq:grpomain}
\tiny
\begin{aligned}
\mathcal{L}(\theta)
&=
\mathbb{E}_{\ell_i \sim \pi_\theta}\!\Big[
\min \Big(
\Gamma_\theta(\ell_i) A(\ell_i),
\mathrm{clip}\!\big(
r_\theta(\ell_i), 1-\epsilon, 1+\epsilon
\big) A(\ell_i)
\Big) \\
&\quad - \beta\, \mathrm{KL}\!\left(\frac{\pi_{\theta}}{\pi_{\theta_{\mathrm{ref}}}}\right)\Big]
\end{aligned}
\vspace{-0.5em}
\end{equation}where 
{\small$\Gamma_\theta(\ell_i)
=
\frac{\pi_\theta(\ell_i)}{\pi_{\theta_{\mathrm{old}}}(\ell_i)}$}. 
This group-relative update prioritizes higher-reward layouts while stabilizing learning across diverse candidates.

\section{Experiments}
\label{sec:exp}
\vspace{-0.7em}
Our experiments address three questions:
(Q1) Does explicit reward-driven policy optimization enable \lays to learn spatial reasoning beyond prompt-based LLM baselines?
(Q2) How does \lays compare with autoregressive and multimodal layout generators in structural validity and visual quality?
(Q3) What performance gap remains between text-based, policy-optimized layout design and specialized task-specific generators?

\vspace{-0.6em}
\paragraph{Datasets} We evaluate on CGL \cite{zhou2022composition} and PKU-PosterLayout \cite{hsu2023posterlayout}. PKU contains text, logo, and underlay elements, while CGL additionally includes embellishments (More details are on Appendix \ref{sec:dataset}). We train on 3,000 randomly sampled layouts per dataset and evaluate on full 1000 CGL and 905 PKU test instances. 
\vspace{-0.6em}
\paragraph{Experimental Setup} \lays uses Qwen-2.5-7B-Instruct \cite{qwen2.5} as the backbone, fine-tuned via LoRA \cite{hu2022lora}. Salient regions are detected using off-the-shelf models \cite{qin2019basnet} and converted to bounding boxes. Element geometries are masked with a \texttt{[MASK]} token, prompting the LLM to predict their positions $(x_{i}, y_{i})$ and sizes $(w_{i}, h_{i})$.
 \vspace{-0.6em}
\paragraph{Baselines} We evaluate \lays against: (i) autoregressive DS-GAN \citep{hsu2023posterlayout}, (ii) VLM-based PosterLlama \cite{seol2024posterllama}, (iii) Qwen-2.5-7B-Instruct in a zero-shot setting, and (iv) proprietary GPT-4o \cite{hurst2024gpt}. (More details in Appendix.\ref{app:baseline})

 \vspace{-0.6em}
\paragraph{Evaluation Metrics}
We evaluate \lays along two dimensions: (1) improvement over the base LLM, using structural metrics including collision, alignment, spacing, and distribution consistency to assess the effectiveness of reward-guided policy learning; and (2) performance against SOTA layout methods, using graphic- and content-focused metrics \cite{hsu2023posterlayout, horita2024retrieval}: Overlay (Ove$\downarrow$) measures overlap among non-underlay elements, Underlay Effectiveness (Und $\uparrow$) assesses how well underlays support content, and Occlusion (Occ $\downarrow$) quantifies overlap with salient regions.

\section{Results and Analysis}
\label{sec:results}
\vspace{-0.6em}
\subsection{Quantitative Results}
\label{sec:quant_eval}
\begin{figure}[ht]
    \vspace{-0.5em}
    \centering
    \begin{subfigure}[t]{0.23\textwidth}
        \centering
        \includegraphics[width=\textwidth]{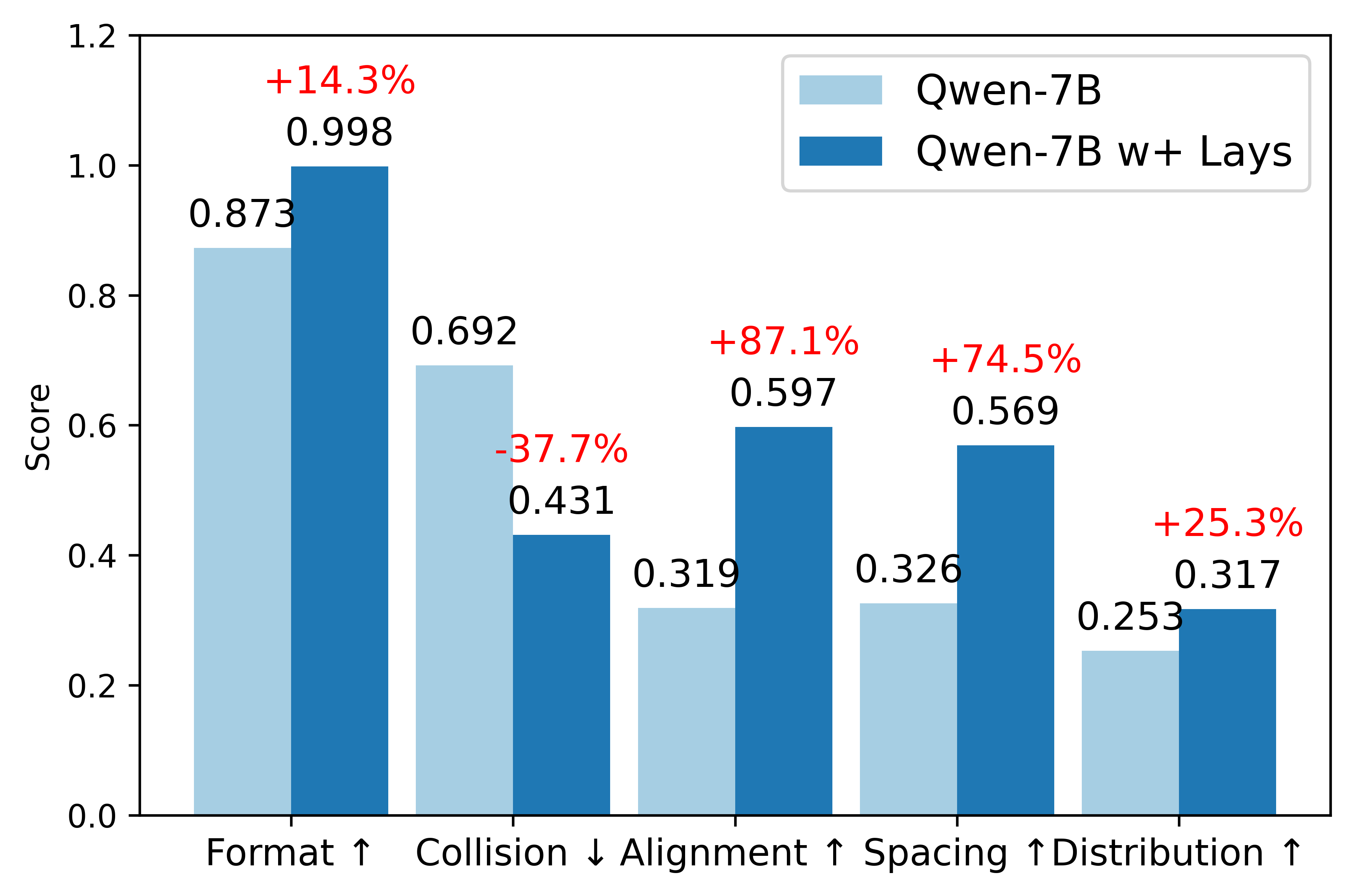}
        \caption{CGL Dataset}
        \label{fig:CGL}
    \end{subfigure}%
    \hspace{0.01\textwidth}%
    \begin{subfigure}[t]{0.23\textwidth}
        \centering
        \includegraphics[width=\textwidth]{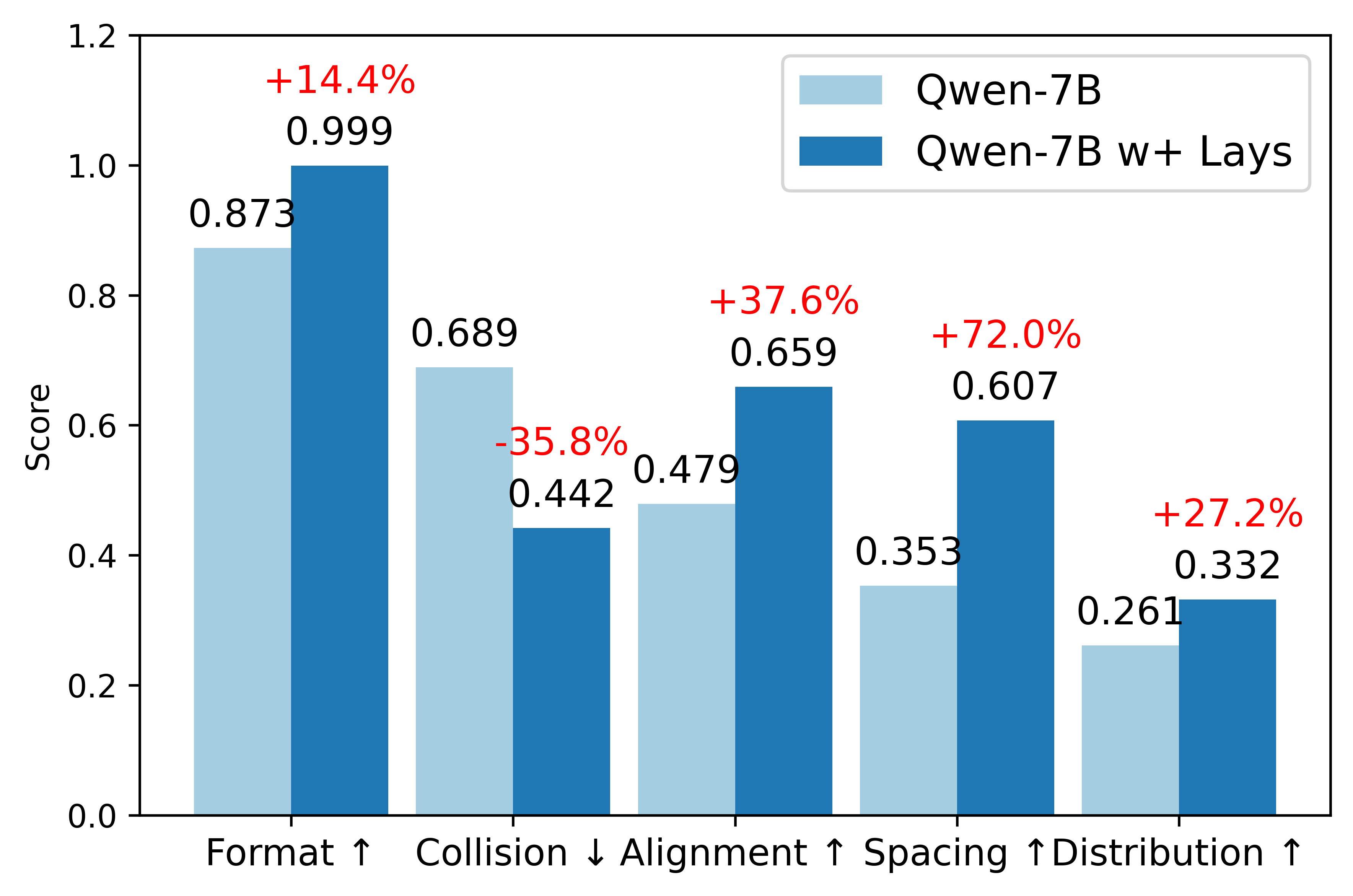}
        \caption{PKU Dataset}
        \label{fig:PKU}
    \end{subfigure}
    \vspace{-0.5em}
    \caption{Comparison of Qwen-7B w/wo \lays on CGL and PKU. Red highlights improvements.}
    \label{fig:learning_effectiveness}
\vspace{-0.5em}
\end{figure}

\vspace{-0.3em}
\paragraph{Learning Effectiveness} 
Figure~\ref{fig:learning_effectiveness} compares the base Qwen model with its \lays-fine-tuned counterpart on the CGL and PKU benchmarks. \lays delivers consistent and substantial gains across all metrics, including alignment ($+63\%$), spacing consistency ($+73\%$), format correctness ($+14\%$) and distribution consistency ($+26\%$), together with a reduction in collision rate ($-36\%$). These gains indicate that reward-driven optimization enables the model to internalize explicit geometric and relational constraints, yielding layouts that better satisfy structural requirements. The largest gains obtained in alignment and spacing suggest that \lays particularly strengthens global spatial organization and inter-element uniformity.

\vspace{-1em}
\paragraph{Compared with SOTA} Table~\ref{tab:sota} reports results on CGL and PKU by different methods. PosterLlama achieves the strongest overall performance, benefiting from its visual encoders and task-specific architectural priors tailored to layout generation. \lays ranks second, substantially improving over its base model, with pronounced reductions in overlap and occlusion and improved underlay effectiveness, indicating more accurate functional grouping and spatial allocation. Despite larger scale, GPT-4o performs moderately, suggesting that model size alone does not guarantee fine-grained spatial reasoning in layout tasks. Notably, \lays attains competitive performance using only 3000 training samples per dataset, far fewer than vision-centric methods, demonstrating the strong spatial inductive bias and sample efficiency induced by reward-driven policy optimization.

\vspace{-0.3em}
\begin{table}[h]
\centering
\scriptsize % smaller font to fit single column
\setlength{\tabcolsep}{3pt} % reduce horizontal padding
\vspace{-0.8em}
\caption{Comparison of overlap (Ove↓), underlay effectiveness (Und↑) and occlusion (Occ↓) on CGL and PKU datasets. \textcolor{gray}{``Real Data"} denotes the annotated layout. Best results are \textbf{bolded}; second-best are \underline{underlined}.}
\vspace{-0.3em}
\setlength{\tabcolsep}{2pt}
\begin{tabular}{c c c c c c c c}
\toprule
 & \#Params & \multicolumn{3}{c}{CGL Dataset} & \multicolumn{3}{c}{PKU Dataset} \\
\cmidrule(lr){3-5} \cmidrule(lr){6-8}
Model & & Ove ↓ & Und ↑ & Occ ↓ & Ove ↓ & Und ↑ & Occ ↓ \\
\midrule
\textcolor{gray}{Real Data} & - & \textcolor{gray}{0.0003} & \textcolor{gray}{0.9926} & \textcolor{gray}{0.1379} & \textcolor{gray}{0.0013} & \textcolor{gray}{0.9974} & \textcolor{gray}{0.1828} \\
DS-GAN & 30M & 0.0361 & 0.6309 & 0.1521 & 0.0336 & 0.7613 & 0.2574 \\
PosterLLama & 7B & \textbf{0.0024} & \textbf{0.9918} & \textbf{0.1476} & \textbf{0.0032} & \textbf{0.9998} & \underline{0.2087} \\
GPT-4o & 200B & 0.0365 & 0.5873 & 0.1591 & 0.0371 & 0.6384 & 0.2743 \\
Qwen-7B & 7B & 0.0474 & 0.5729 & 0.1615 & 0.0479 & 0.6059 & 0.2384 \\
Qwen-7B$+$\lays & 7B & \underline{0.0257} & \underline{0.6989} & \underline{0.1487} & \underline{0.0260} & \underline{0.7688} & \textbf{0.2072} \\
\bottomrule
\end{tabular}

\label{tab:sota}
\end{table}

 \vspace{-1.5em}
\subsection{Qualitative Results}
\label{sec:qual}
\begin{figure}[htbp]
  \centering
  \subfloat[CGL\label{fig:cgl}]{
    \includegraphics[width=0.502\columnwidth]{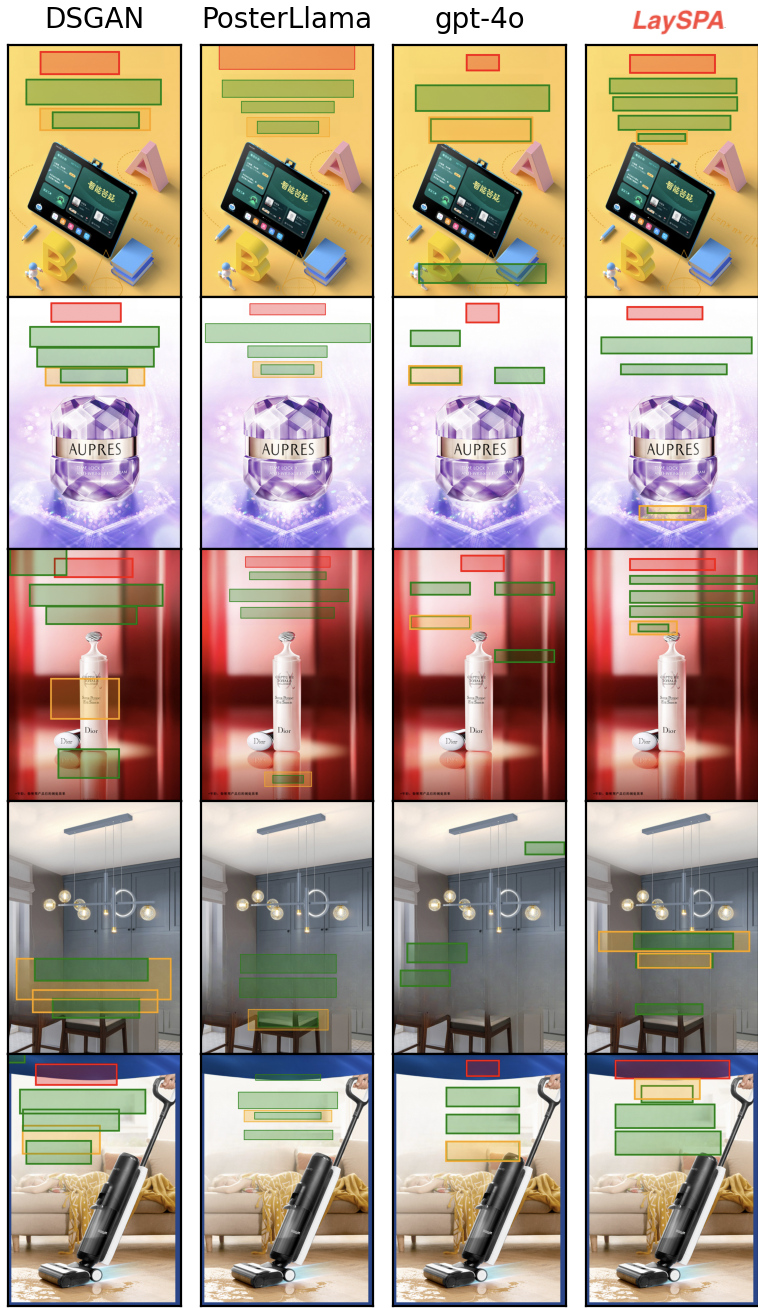}
  }
  \subfloat[PKU\label{fig:pku}]{
    \includegraphics[width=0.505\columnwidth]{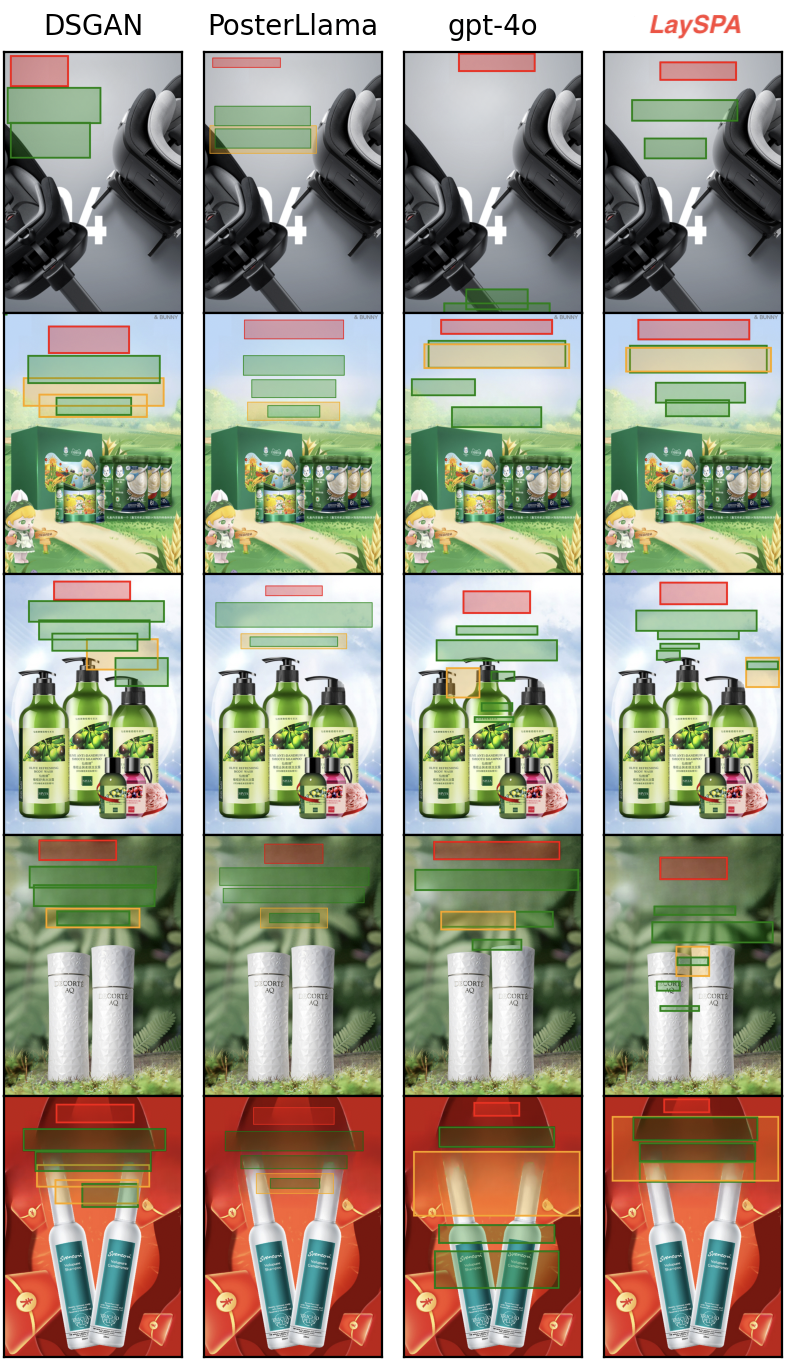}
  }
 \vspace{-1em}
  \caption{Layouts generated by different methods.}
  \label{fig:combined}
   \vspace{-1em}
\end{figure}

\vspace{-0.3em}
Figure~\ref{fig:combined} presents qualitative comparisons across methods. DS-GAN achieves reasonable canvas coverage and consistent element sizing but often exhibits misalignment, unintended overlaps, and interference with salient regions, reflecting limited global spatial coordination. PosterLlama produces the most visually coherent layouts, benefiting from vision-specific encoders that impose strong structural priors and balanced compositions. GPT-4o underperforms particularly on CGL, with frequent misalignment and weak relational organization, revealing limitations in fine-grained spatial control despite its scale. In contrast, \lays consistently generates visually harmonious and well-proportioned layouts, demonstrating that explicitly learned spatial reasoning and policy optimization improves overall design quality.
 
\vspace{-0.6em}

\subsection{Ablation Study}
\vspace{-0.3em}
To dissect the contributions of each component in the hybrid reward (Eq.~\ref{eq:reward}), we vary $(\lambda_f, \lambda_q, \lambda_u)$ and evaluate \lays on the CGL benchmark. We consider four configurations: (1) Format-focused $(0.5, 0.4, 0.1)$ emphasizes schema compliance, (2) Quality-focused $(0.1, 0.8, 0.1)$ prioritizes geometric and visual design principles, (3) IoU-focused $(0.1, 0.1, 0.8)$  emphasizes alignment with reference layouts, and (4) Balanced-hybrid $(0.1, 0.45, 0.45)$ distributes weight across all objectives.

\begin{figure}[htb]
\vspace{-0.5em}
\centering
\includegraphics[width=0.8\columnwidth]{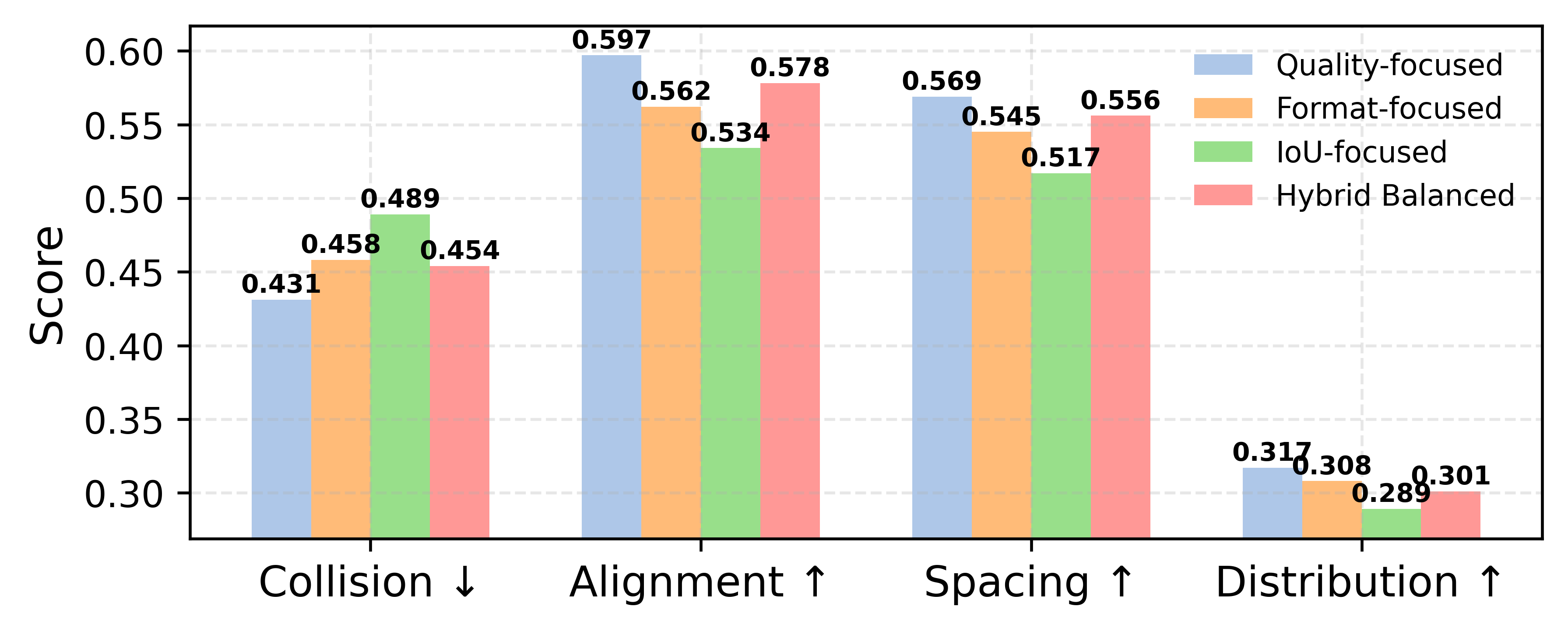} 
\vspace{-0.5em}
\caption{Ablation study on varying reward weights}
\label{ablation}
 \vspace{-1em}
\end{figure}

Figure~\ref{ablation} shows performance across collision, alignment, spacing consistency, and distribution metrics. Quality-focused training yields the best overall results, highlighting the importance of explicit geometric and relational objectives for coherent, functionally valid layouts. The balanced-hybrid setting offers a pragmatic trade-off, maintaining reasonable reference alignment while preserving structural fidelity. In contrast, IoU-focused underperforms, indicating that optimizing for reference matching alone leads to brittle spatial reasoning and insufficient attention to geometric constraints.
\vspace{-1em}
\section{Related Works}
\label{sec:related}
\vspace{-0.5em}
\paragraph{Spatial Reasoning with LLMs}
Vision–Language Models (VLMs) \citep{fan2025vlm} and Multimodal LLMs (MLLMs) \citep{wu2025spatial} enable spatial reasoning by grounding language in visual inputs, using multi-view images \citep{qi12shapellm, 10943341, gu2024conceptgraphs}, depth cues \citep{ma2024spatialpin, cheng2024spatialrgpt, cai2025spatialbot} or cross-modal embeddings \citep{luddecke2022image, chen2024spatialvlm}. While effective, these methods are computationally intensive, encode geometry implicitly, and offer limited interpretability. 
\vspace{-1em}
\paragraph{Content-Aware Layout Generation}
Traditional layout generation relies on deep generative models such as VAEs \citep{jyothi2019layoutvae}, GANs \citep{zhou2022composition, hsu2023posterlayout, li2019layoutgan}, transformers \citep{kong2022blt}, autoregressive \citep{horita2024retrieval}, and diffusion-based models \citep{chai2023two, chai2023layoutdm}, supplemented by retrieval \citep{horita2024retrieval} or post-hoc refiners \citep{shen2025layoutrectifier} to reduce overlap and misalignment. These approaches rely heavily on annotated data and thus generalizing poorly beyond training distributions. \lays differs by formulating layout design as reinforcement learning over a structured textual environment, enabling diverse exploration of geometrically valid layouts guided by spatial objectives.
\vspace{-2em}
\paragraph{LLM-based Layout Generation} 
Recent works leverage LLMs’ design priors by serializing layouts into text formats such as HTML \citep{tang2023layoutnuwa, seol2024posterllama}, CSS \citep{feng2023layoutgpt}, or SVG \citep{hsu2025postero} and applying recursive reasoning to capture spatial dependencies \citep{tian2025relayout}. Retrieval-based pipelines \citep{lin2023layoutprompter, forouzandehmehr2025cal} ground outputs in exemplar layouts and iteratively refine them using LLMs and vision–language feedback. These methods often rely on emergent capabilities and exemplar retrieval, whereas \lays is explicitly guided by spatial and visual rewards and optimized through peer-relative policy learning.

\section{Conclusion}
\label{sec:conclusion}
\vspace{-0.6em}
We introduce \lays, a GRPO-based framework that reframes content-aware layout design as a policy learning problem, guiding LLMs to perform explicit spatial reasoning over structured textual environments. Optimized with multi-objective spatial critiques, \lays learns robust and generalizable spatial behaviors for layout design. Empirically, \lays substantially improves layout quality, outperforms larger proprietary LLMs, and approaches specialized multimodal generators without visual inputs and less training samples. These results demonstrate that spatial reasoning can be explicitly optimized in text, positioning language-grounded, reward-driven policy learning as a scalable alternative to vision-centric layout design. Limitations include reliance on precomputed saliency and single-step generation. Future work will explore richer semantic grounding, multi-turn reinforcement learning, and hierarchical design policies.

\iffalse
\section*{Limitations}

This document does not cover the content requirements for ACL or any
other specific venue.  Check the author instructions for
information on
maximum page lengths, the required ``Limitations'' section,
and so on.
\fi

\section*{Acknowledgments}
Original images from PKU-PosterLayout dataset by Hsu et al. available at https://huggingface.co/datasets/creative-graphic-design/PKU-PosterLayout, licensed under CC BY-SA 4.0. Original images from CGL dataset by Zhou et al. available at https://huggingface.co/datasets/creative-graphic-design/CGL-Dataset, licensed under CC BY-NC-SA 4.0.

\bibliography{paper}

\appendix

\section{Visualization of Rewards}
\label{app:visreward}
\begin{figure}[htbp]
\centering
\renewcommand{\arraystretch}{1.2} % adjust vertical spacing
\setlength{\tabcolsep}{3pt}      % adjust column spacing
\begin{tabular}{m{0.2\linewidth}|m{0.35\linewidth}|m{0.35\linewidth}}

\centering Collision $R_{\tiny{icr}}$ &
\includegraphics[width=0.99\linewidth]{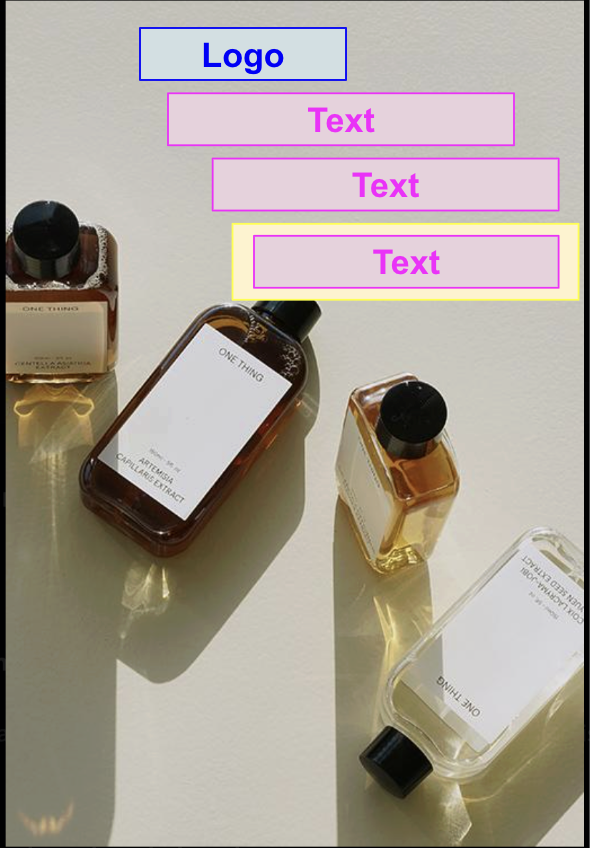} &
\includegraphics[width=0.99\linewidth]{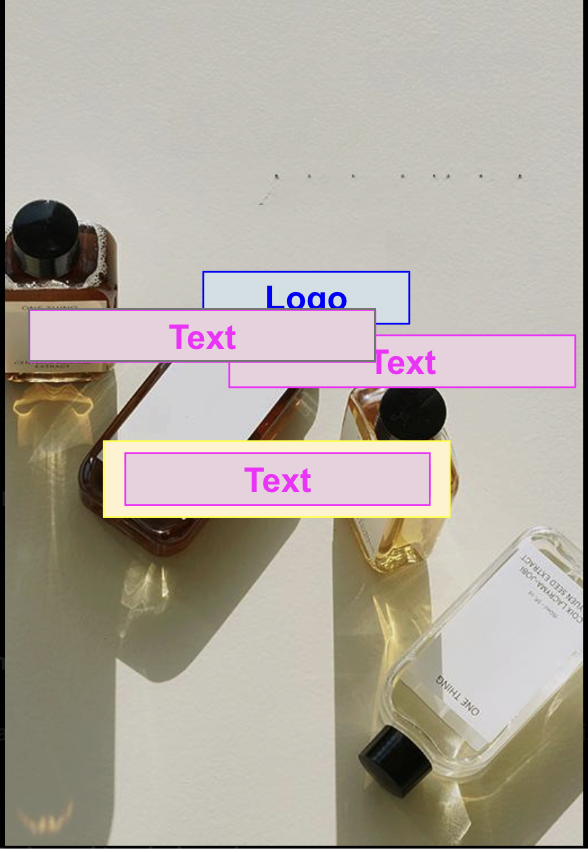} \\
\hline
\centering Alignment $R_{\tiny{al}}$ &
\includegraphics[width=0.99\linewidth]{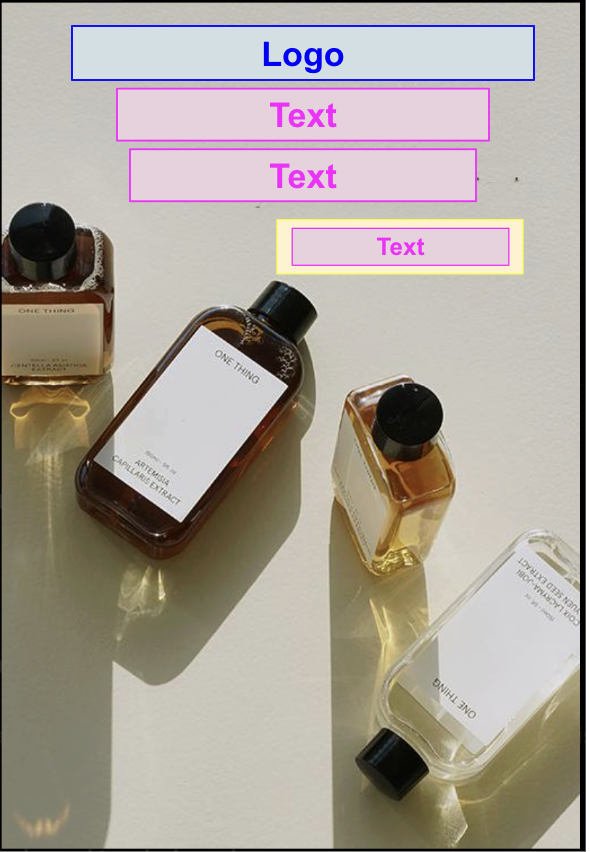} &
\includegraphics[width=0.99\linewidth]{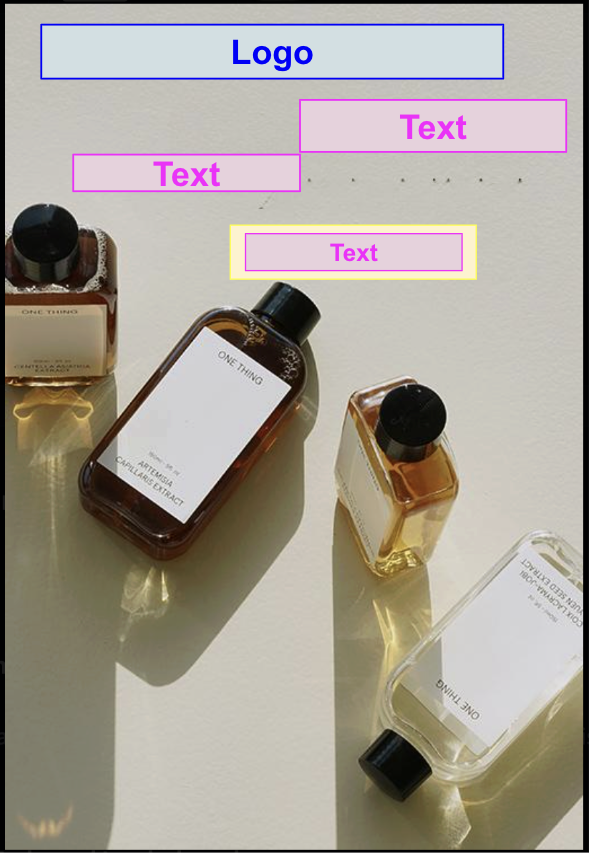} \\
\hline
\centering Distribution $R_{\tiny{dis}}$ &
\includegraphics[width=0.99\linewidth]{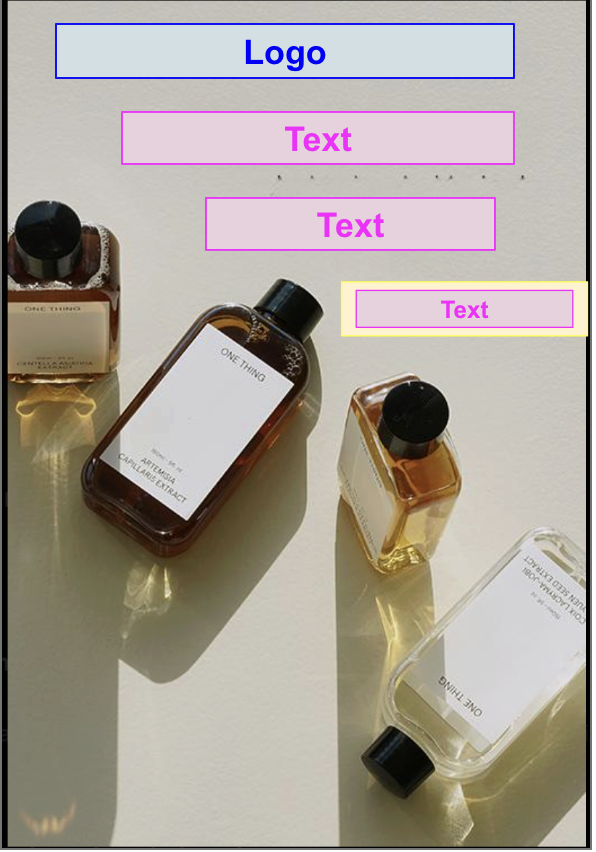} &
\includegraphics[width=0.99\linewidth]{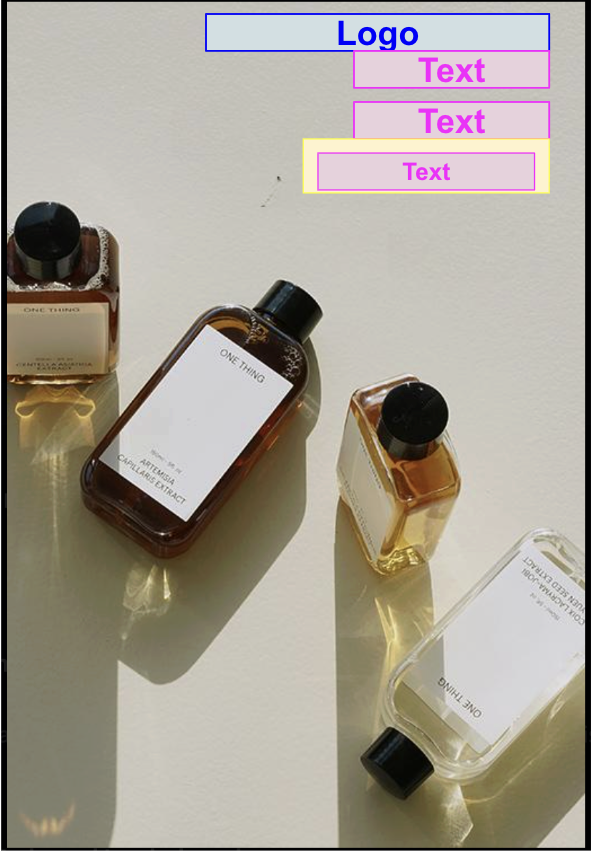} \\
\hline
\centering Spacing $R_{\tiny{sp}}$ &
\includegraphics[width=0.99\linewidth]{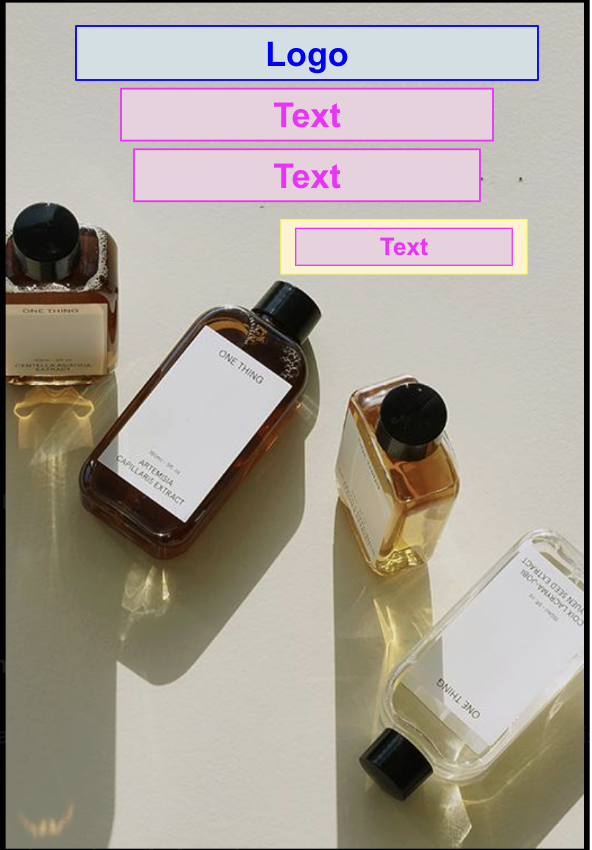} &
\includegraphics[width=0.99\linewidth]{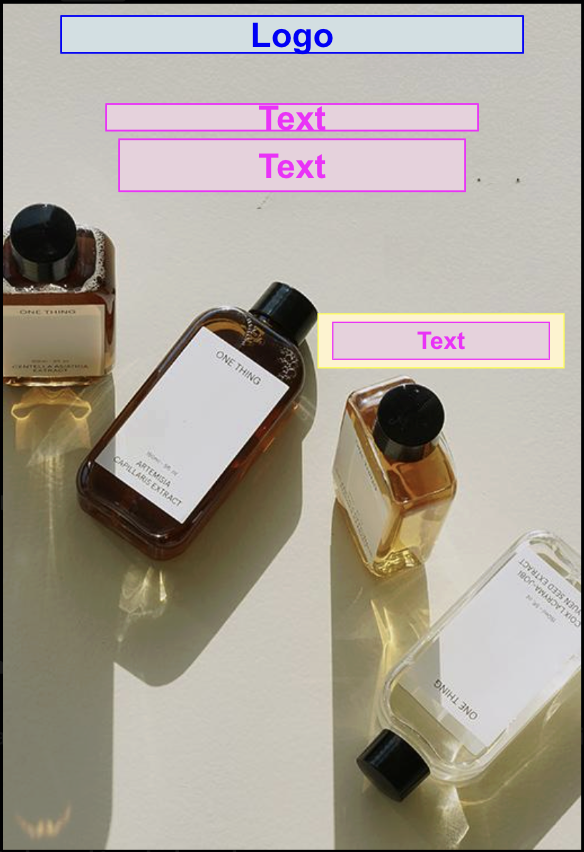} \\
\hline
\centering Underlay-text $R_{\tiny{ut}}$ &
\includegraphics[width=0.99\linewidth]{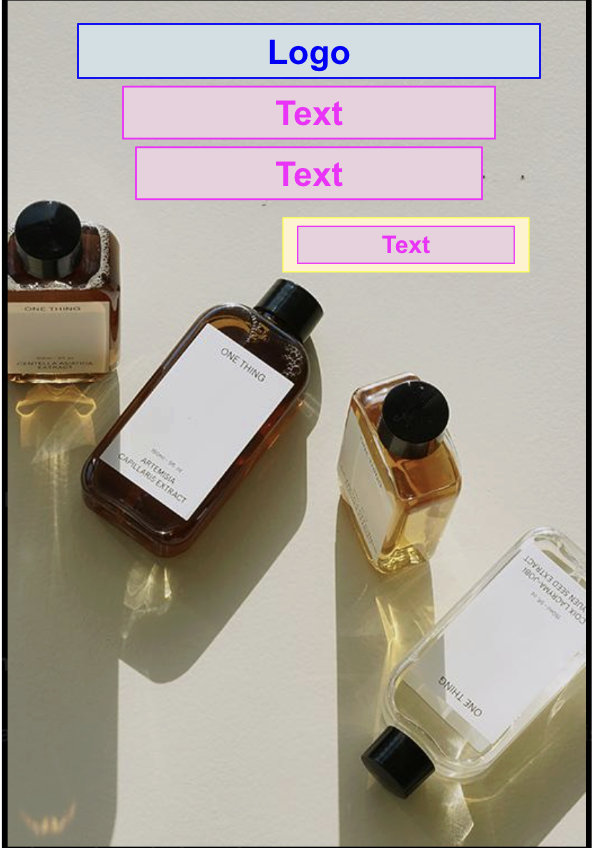} &
\includegraphics[width=0.99\linewidth]{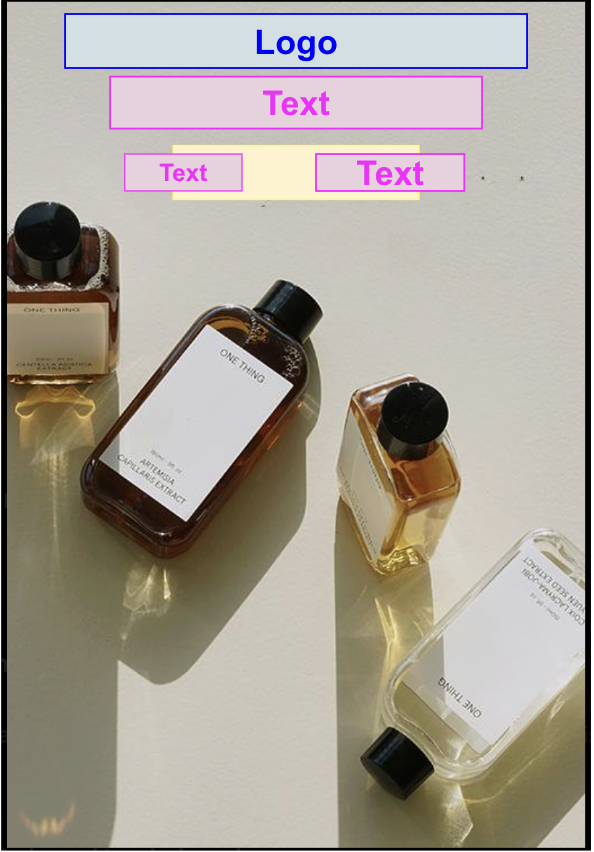} \\
\hline
 &
\centering \textbf{\textcolor{Green}{\ding{52}}} &
\centering \textbf{\textcolor{red}{\ding{56}}}\\

\end{tabular}
\caption{An illustration of how each layout quality score functions guides the agent toward human-preferred designs (\textbf{\textcolor{Green}{\ding{52}}}) while discouraging (\textbf{\textcolor{red}{\ding{56}}}) undesirable spatial arrangements.
}
\label{fig:textimagegrid}
\end{figure}

Figure~\ref{fig:textimagegrid} illustrates how individual components of the layout quality score guide the agent toward content-aware designs that jointly satisfy structural feasibility and visual appeal.

\section{Dataset}
\label{sec:dataset}
Table \ref{tab:dataset} lists the details of the CGL and PKU benchmark.

\begin{table}[htb]
\centering
\caption{Layout generation datasets.}
\footnotesize
\setlength{\tabcolsep}{2.5pt}
\renewcommand{\arraystretch}{1.05}
\begin{tabular}{|c|c|c|}
\hline
\textbf{Dataset} 
& \makecell{\textbf{CGL}} 
& \makecell{\textbf{PKU}} \\
\hline
\textbf{\# Layouts} 
& 60{,}548 
& 9{,}974 \\
\hline
\textbf{\# Canvases} 
& 1{,}000 
& 905 \\
\hline
\textbf{Elements} 
& \makecell{Text, Logo, \\ Underlay, Emb.}
& \makecell{Text, Logo, \\ Underlay} \\
\hline
\textbf{Canvas} 
& Non-empty 
& Non-empty \\
\hline
\textbf{Categories} 
& \makecell{Cosmetics, Elec., \\ Clothing, etc.}
& \makecell{Cosmetics, Elec., Clothing, \\ Food, Toys, etc.} \\
\hline
\end{tabular}
\label{tab:dataset}
\end{table}

\iffalse
\begin{table}[htb]
\centering
\caption{Summary of layout generation datasets.}
\setlength{\tabcolsep}{4pt}
\renewcommand{\arraystretch}{1.15}
\begin{tabular}{|c|c|c|p{3.2cm}|c|p{4.2cm}|}
\hline
\multirow{2}{*}{Dataset}
& \multicolumn{2}{c|}{\# Samples}
& \multirow{2}{*}{Element Types}
& \multirow{2}{*}{Canvas}
& \multirow{2}{*}{Content Category} \\
\cline{2-3}
& Layout & Canvas &  &  &  \\
\hline
\makecell{CGL \\ \citep{zhou2022composition}}
& 60,548
& 1,000
& Text, logo, underlay, embellishment
& Non-empty
& Cosmetics, electronics, clothing, etc. \\
\hline
\makecell{PKU PosterLayout \\ \citep{hsu2023posterlayout}}
& 9,974
& 905
& Text, logo, underlay
& Non-empty
& Cosmetics, electronics, clothing, delicatessen, toys/instruments, etc. \\
\hline
\end{tabular}
\label{tab:dataset}
\end{table}
\fi
\section{Baseline Methods}
\label{app:baseline}
\begin{itemize}
    \item \textbf{DS-GAN}~\citep{hsu2023posterlayout}: DS-GAN is a CNN-LSTM-based conditional generative adversarial network that learns design sequences conditioned on a given canvas to automatically generate content-aware visual-textual presentation layouts.
    \item \textbf{PosterLlama}~\citep{seol2024posterllama}: a multimodal layout generation system that uses CodeLLaMA-7B as the language backbone with a DINOv2 visual encoder. Layouts are reformulated as HTML sequences.
The model is trained via supervised instruction tuning with depth-based data augmentation.

\item \textbf{Qwen-2.5-7B-Instruct}: A dense transformer language model with 7 billion parameters. Given the same inputs as \lays, while used in a training-free, prompt-based setting to generate geometric attributes for all layout elements.
\item \textbf{GPT-4o}: A large-scale proprietary multimodal foundation model that takes the canvas as visual input and is prompted to generate the geometric specifications of all layout elements.
\end{itemize}

\end{document}